# A Deep-Learning-Based Fashion Attributes Detection Model

**Menglin Jia**  **Yichen Zhou**  **Mengyun Shi**  **Bharath Hariharan**
Cornell University
{mj493, yz888, ms2979}@cornell.edu, harathh@cs.cornell.edu

## 1  Introduction

Visual analysis of clothings is a topic that has received increasing attention in computer vision communities recent years. There is already a large body of research on clothing modeling, recognition, parsing, retrieval, and recommendations (See Section 2). Figure S1 summarizes related works since 2010 in terms of purposes and domains. An increasing number of the papers focused on image retrieval on daily life and online websites, a task essential in fashion e-commerce for consumers. Yet little attention has been made on fashion analysis for people who work in fashion industry.

Analyzing fashion attributes is essential in fashion design process. Current fashion forecasting firms, such as WGSN utilizes information from all around the world (from fashion shows, visual merchandising, blogs, etc) [1–3]. They gather information by experience, by observation, by media scan, by interviews, and by exposed to new things. Such information analyzing process is called abstracting, which recognize similarities or differences across all the garments and collections. In fact, such abstraction ability is useful in many fashion careers with different purposes [4]. Fashion forecasters abstract across design collections and across time to identify fashion change and directions; designers, product developers and buyers abstract across group of garments and collections to develop a cohesive and visually appeal lines; sales and marketing executives abstract across product line each season to recognize selling points; fashion journalist and bloggers abstract across runway photos to recognize symbolic core concepts that can be translated into editorial features [5].

Fashion attributes analysis for such "fashion insiders" requires much detailed and in-depth attributes annotation than that for consumers, and requires inference on multiple domains. In this project [1], we propose a data-driven approach for recognizing fashion attributes. Specifically, a modified version of Faster R-CNN model will be trained on images from a large-scale localization dataset with 594 fine-grained attributes under different scenarios, for example in online stores and street snapshots. This model will then be used to detect garment items and classify clothing attributes for runway photos and fashion illustrations.

## 2  Related Work

### 2.1  Detecting clothing categories and attributes

Earlier works adopted non-convolutional-neural-network (CNN) approaches for clothing detection. [6–9] used feature extractors such as SIFT and HOG to classify apparels and described attributes. [10–14] dedicated to clothing segmentations for different categories via probabilistic methods. Some of above preprocessed images with pose estimation or upper/lower body detection. In addition, works like [15, 16] focused on retrieve images that have high clothing similarity based on visual attributes. Later, many works took the CNN way. FashionNet [17] adopted an CNN architecture similar to VGG-16 to predict categories and attributes. [18] tackled segmentation task with a fully-convolutional neural network (FCN) approach. [19–21] utilized R-CNN models for body detection or to generate

---
[1]This work was done as a class project for 'CS6670 Computer Vision' at Cornell University

clothing proposals. Our approach is mostly based on Faster R-CNN to produce clothing proposals and category classifications, and we add extra branches for attribute detection.

### 2.2 Cross-Domain Detection

There have been a number of works tackling the issue of cross-domain clothing detection. The most popular topic is to retrieve similar fashion images from different domains [22–29], many based on deep neural network [24–29]. Most of the works in this area have focused on learning a transformation that aligns the source and target domain representations into a common feature space, or dealing with the cross domain problem with limited amount of labeled datasets available in the target domain [19–21]. We will test our model on different domains, however, we haven't try to tackle this issue yet.

### 2.3 Analyzing Fashion Trend Based on Visual Attributes

Early work on trend analysis [30] broke down catwalk images from NYC fashion shows to find style trends in high-end fashion. Recent advance in deep learning enabled more work on this area. [31–33] utilized deep networks to extract clothing attributes from images and created a visual embedding of clothing style cluster in order to study the fashion trends in clothing around the globe.

### 2.4 Fashion Category and Attribute Dataset

Table 1 shows a comparison among different datasets with clothing category and attribute labels. We use DeepFashion (category and attribute prediction benchmark) [17] for our model since it contains enough number of images, is rich in attribute classes, and has two popular domains.

| Name | # of Images | # of Attributes | Domains |
|---|---|---|---|
| Runway2Realway [15] | 348,598 | 53 | Runway |
| Exact Street2Shop [16] | 78,958 | 11 | Shop + Consumer |
| DARN [21] | 182,780 | 179 | Shop + Consumer |
| StreetStyle-27K [33] | 27,000 | 39 | Consumer |
| DeepFashion (Category and Attribute) [17] | 289,222 | 1050 | Shop + Consumer |
| DeepFashion (In-shop) [17] | 52,712 | 513 | Shop + Consumer |
| DeepFashion (Consumer-to-shop) [17] | 239,577 | 353 | Shop + Consumer |

Table 1: Fashion Dataset Comparison

## 3 Dataset Modification and Model Structure

### 3.1 Dataset Modification

The dataset we choose, DeepFashion (category and attribute prediction benchmark), has some labeling problems. Bounding-box-wise, we removed bounding boxes with odd aspect ratios (height/weight lower than 0.2 or higher than 5) or extremely small area (less than 0.21% of the whole image). Attribute-wise, we manually removed 45 unclear attributes (such as "girl" and "please") and merged semantically similar attributes, (for example, "abstract geo" vs "abstract geo print" vs "geo" vs "geo pattern" vs "geo print"). The cleaned dataset ended up with 544 diverse clothing attributes and 50 clothing categories. There are still wrongly labeled (false positive) categories, attributes and bounding boxes in this dataset, e.g., recognizing a skirt as a dress, but we are not able to deal with them.

### 3.2 Model Architecture

We extend the Faster R-CNN object detection framework [34] with ResNet 101 and ROI-align (implemented by Google Research [35]) with two modifications: a pruning mechanism and additional clothing attributes branches parallel to category branch. Figure S2 shows the overall model architecture.

**Additional pruning.** Faster R-CNN classifies objectiveness of each densely distributed proposals at the first Region Proposal Network (RPN) stage. Each proposal is labeled as positive/negative or ignored based on its IoU value with the groundtruth boxes. Since DeepFashion dataset gives only one box for a single image, other clothing items in this image (especially upper body vs. lower body) will be classified as background. This would confuse the detection model and decrease the performance. In order to solve such problem, we propose an additional pruning process at the first stage. Specifically, we introduce groundtruth people boxes for each image, and prune away any proposals that classified as non-objects but have an IoU of a certain value or higher with groundtruth people boxes (Figure 1). We use SSD-Mobilenet [35] to extract groundtruth people boxes. It is worth mentioning that a small fraction (9.9%) of images display clothes other than models, out of which a large portion display single clothing item.

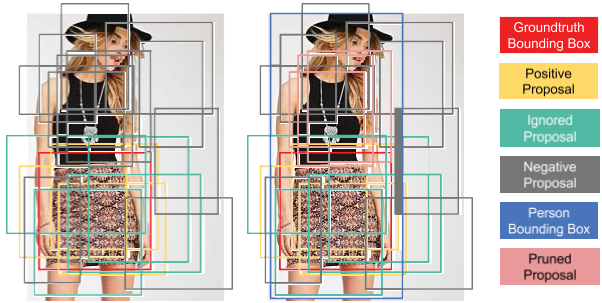

Figure 1: Pruning Mechanism During Training

**Attribute branches.** We approach learning both attributes and category as a multi-task learning problem. The attributes branches reuse features extracted by RPN. We propose three different structures as indicated in Figure 2. In detail, (a) uses 3 convolution layers ($1 \times 1 \times 512$ with padding, $7 \times 7 \times 512$ without padding, and $1 \times 1 \times 2048$ without padding) followed by 5 fully connected layers for each attribute type scores; (b) shares the same flattened proposal features as category classifier, followed by a fully connected layers (1024) and 5 fully connected layers for each attribute type scores; (c) shares the same flattened proposal features as category classifier, followed by 5 fully connected layers for each attribute type scores.

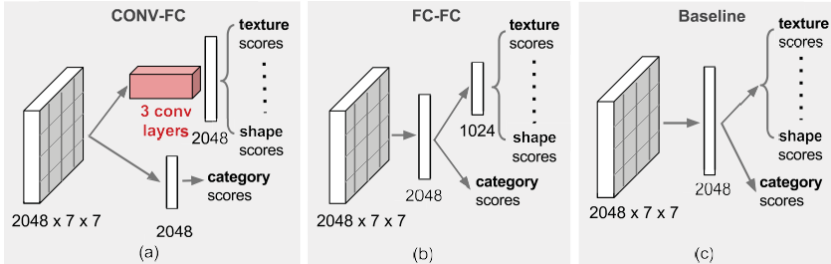

Figure 2: Attribute Branches

## 4 Experiment and Analysis

### 4.1 Test and Results

We test our trained model on three datasets: (a) $2,094$ selected images from DeepFashion (category and attribute prediction benchmark) test partition; (b) 92 images from ready-to-wear runway photos; (c) 92 images from fashion technical sketches. By default, 300 detections are generated for each image without score threshold. For evaluation of category prediction, we consider two metrics: **(i) average precision (AP) per class and weighted mAP.** We use all the predictions with scores higher than 0.5 per image to compute true-positive and false-positive labels per class with a matching IoU threshold of 0.5 with groundtruth boxes. Then these labels are used to calculate AP per class and weighted mAP (concatenating all labels regardless of classes); **(ii) CorLoc per class and weighted mean CorLoc.** For each image, we pick the top 5 detections per image and check if the grountruth is correctly detected per class with a matching IoU threshold of 0.5 with groundtruth boxes. CorLoc is computed as the ratio of number of detected groundtruth instances over number of total groundtruth instances per class, and weighted mean CorLoc is such ratio regardless of classes. For evaluation of attribute prediction (only on DeepFashion dataset), we label attribute detections as positive and negative with a score threshold of 0.5. For each image, we merge all detections that have an IoA higher than 0.7 over the detection with the highest category score, using logical "and" operation. Then we calculate true-positive and false-positive labels for each attribute and generate **precision and recall per attribute and precision and recall per attribute type**. Table 2 presents the corresponding test results. We used two IoU threshold (0.3 and 0.7) for additional pruning and three structures as attribute branch. Comparisons are made with models restored from checkpoints under the same epoch. Note that conv-fc attribute branch doesn't give any positive attribute prediction.

| Dataset | weighted mAP | | | weighted mean CorLoc | | |
|---|---|---|---|---|---|---|
| | no pruning | pruning 0.3 | pruning 0.7 | no pruning | pruning 0.3 | pruning 0.7 |
| DeepFashion | 0.1425 | 0.1603 | 0.1715 | 0.6418 | 0.6336 | 0.6772 |
| Runway | 0.0996 | 0.0865 | 0.0980 | 0.4130 | 0.4130 | 0.4891 |
| Sketches | 0.0639 | 0.0422 | 0.0748 | 0.3804 | 0.3261 | 0.3913 |

| Dataset | Attribute Precision | | | Attribute Recall | | |
|---|---|---|---|---|---|---|
| | baseline | fc-fc | conv-fc | baseline | fc-fc | conv-fc |
| DeepFashion | 0.0448 | 0.0899 | - | 0.2932 | 0.2127 | - |

Table 2: Test Results

## 4.1 Analysis

From the experimental results, we can see that our model doesn't give a great performance. Here are some factors which we think have negative effects.

**Data imbalance.** The training data we used consists of 46 categories and some categories have only tens or hundreds of images while other categories can have over $70,000$ images. This imbalance makes it really hard to train the model so that it can detect those minor categories.

**Wrongly labeled data.** There are still a lot of wrongly labeled categories and attributes in the dataset even after our data cleaning.

**Unbounded objects.** We tackled this issue using proposal pruning, however, there are still such cases considering the limitation of pruning criterion and they also occur in the test data, which adversely affects the evaluation.

**Too many negative attribute labels.** In the training data, each attribute has way more negative instances than positive ones. We didn't deal with issue, and as a result, the attribute classifier is not well established.

## 5 Future Work

**Optimization methods.** To improve optimization performance, we want to compare different optimization strategies. We'd like to explore using Adam optimizer with manual learning rate decay compare different optimization strategies and using batch size of 1 instead of mini-batch for gradient descent.

**Dealing with data imbalance**: As mentioned above, there's significant data imbalance among categories and between positive and negative labels for attributes. Stratified sampling [33] and weighted cross-entropy loss [17] might be of help.

**Finer feature recognition.** We use feature maps with low resolutions but large receptive field, thus detailed attributes on clothes (such as side-zippers, or small embroideries at collars) may not be easily recognized. We may consider using Feature Pyramid Networks (FPN) or multi-scale DenseNet to improve it.

**Domain adaptation.** We will explore how to improve detection performance of a model trained from one domain on another domains (haute couture runway photos, artistic fashion drawings, etc.).

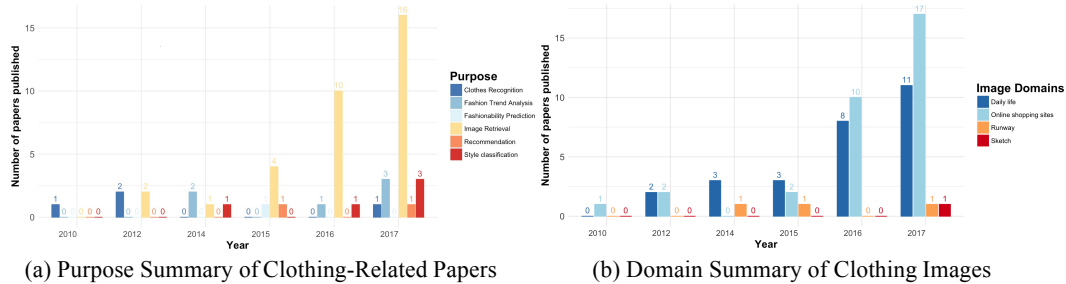

(a) Purpose Summary of Clothing-Related Papers    (b) Domain Summary of Clothing Images

Figure S1: Summary of Clothing Analysis Papers Since 2010

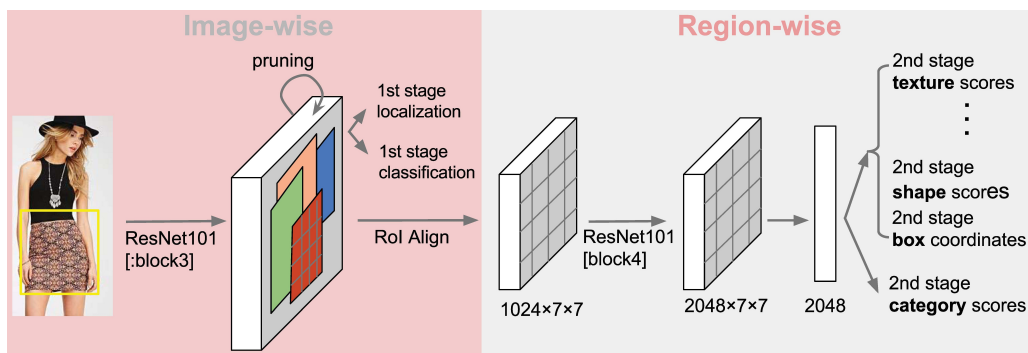

Figure S2: Model Architecture